\documentclass[sigconf]{acmart}

\usepackage{subfigure}
\usepackage{multirow}

\AtBeginDocument{%
  \providecommand\BibTeX{{%
    \normalfont B\kern-0.5em{\scshape i\kern-0.25em b}\kern-0.8em\TeX}}}

\copyrightyear{2023}
\acmYear{2023}
\setcopyright{acmcopyright}\acmConference[WSDM '23]{Proceedings of the Sixteenth ACM International Conference on Web Search and Data Mining}{February 27-March 3, 2023}{Singapore, Singapore}
\acmBooktitle{Proceedings of the Sixteenth ACM International Conference on Web Search and Data Mining (WSDM '23), February 27-March 3, 2023, Singapore, Singapore}
\acmPrice{15.00}
\acmDOI{10.1145/3539597.3570487}
\acmISBN{978-1-4503-9407-9/23/02}

\acmConference[WSDM '23]{Make sure to enter the correct
  conference title from your rights confirmation emai}{February 27-March 3, 2023}{Singapore, Singapore }

\acmPrice{15.00}
\acmISBN{978-1-4503-9407-9/23/02}

\begin{document}

\title{Beyond Digital “Echo Chambers”: The Role of Viewpoint Diversity in Political Discussion}

\author{Rishav Hada}
\email{rishavhada@gmail.com}
\affiliation{%
  \institution{Microsoft Research India}
  \city{Bengaluru}
  \country{India}
}
\author{Amir Ebrahimi Fard}
\email{a.ebrahimifard@maastrichtuniversity.nl}
\affiliation{%
  \institution{Maastricht University}
  \city{Maastricht}
  \country{Netherlands}
  }
\author{Sarah Shugars}
\email{sarah.shugars@rutgers.edu}
\affiliation{%
  \institution{Rutgers University}
  \city{New Brunswick}
  \country{USA}
}
\author{Federico Bianchi}
\email{fede@stanford.edu}
\affiliation{%
  \institution{Stanford University}
  \city{Stanford}
  \country{USA}
  }
  
  \author{Patricia Rossini}
\email{patricia.rossini@glasgow.ac.uk}
\affiliation{%
  \institution{University of Glasgow}
  \city{Glasgow}
  \country{United Kingdom}
}
\author{Dirk Hovy}
\email{dirk.hovy@unibocconi.it}
\affiliation{%
  \institution{Bocconi University}
  \city{Milan}
  \country{Italy}
  }
  
   \author{Rebekah Tromble}
\email{rtromble@gwu.edu}
\affiliation{%
  \institution{ George Washington University}
  \city{Washington, D.C.}
  \country{USA}
}
\author{Nava Tintarev}
\email{n.tintarev@maastrichtuniversity.nl}
\affiliation{%
  \institution{ Maastricht University}
  \city{Maastricht}
  \country{Netherlands}
  }

\renewcommand{\shortauthors}{Hada et al.}

\begin{abstract}
Increasingly taking place in online spaces, modern political conversations are typically perceived to be unproductively affirming---siloed in so called ``echo chambers'' of exclusively like-minded discussants.  
Yet, to date we lack sufficient means to measure viewpoint diversity in conversations. To this end, in this paper, we operationalize two viewpoint metrics proposed for recommender systems and adapt them to the context of social media conversations. This is the first study to apply these two metrics \textit{(Representation and Fragmentation)} to real world data and to consider the implications for online conversations specifically. 
We apply these measures to two topics---daylight savings time (DST), which serves as a control, and the more politically polarized topic of immigration. 
We find that the diversity scores for both Fragmentation and Representation are lower for immigration than for DST. 
Further, we find that while pro-immigrant views receive consistent pushback on the platform, anti-immigrant views largely operate within echo chambers. We observe less severe yet similar patterns for DST.  
Taken together, Representation and Fragmentation paint a meaningful and important new picture of viewpoint diversity. 

\end{abstract}

\ccsdesc[500]{Computing methodologies~Discourse, dialogue and pragmatics}

\keywords{Twitter, conversation network, viewpoint diversity, echo chambers}

\maketitle
\section{Introduction}
Political conversations between everyday people form the foundation of a healthy democracy \cite{mansbridge1999everyday}. In theory, exchanging perspectives allows citizens to collaboratively identify the best solutions to shared problems and builds democratic legitimacy for the implementation of those solutions \cite{cohen1989deliberation,dryzek2009democratization}. In practice, however, there are many reasons to be skeptical that these political conversations are actually achieving their goals. Increasingly taking place in online spaces, modern political conversations are typically perceived to be unproductively affirming---siloed in so called ``echo chambers'' of exclusively like-minded discussants \cite{barbera2015tweeting,bakshy2015exposure}. However, this focus on the worldview or ideology of discussants overlooks a crucial ingredient of discursive democratic theory: the viewpoints expressed in conversation. For democratic discourse to be productive, it is in some ways less important that the interlocutors themselves embrace different ideologies than that they are aware of and come into contact with different views \cite{mercier2012reasoning}. On relatively public and open social media platforms, such as Reddit or Twitter, it may be more likely for those who hold a particular opinion to encounter divergent views as part of comment and reply threads. Yet to date, most research examining echo chambers has focused on the ideology of users and those they follow, rather than the specific viewpoints they engage with or to which they are exposed \cite{bakshy2015exposure,barbera2015birds,barbera2015tweeting,bastos2018geographic,cinelli2021echo}.

The largely understudied dimension of viewpoint diversity serves as the primary focus of this work.  We take inspiration from the viewpoint diversity metrics conceptualized for the news recommender systems domain by \citet{vrijenhoek2021recommenders}. We adapt the definition and propose novel operationalization of two such metrics to measure viewpoint diversity for the domain of social media conversations. 
This is the first study to apply the two metrics to real world data, in this case, online conversations. We also present in-depth analysis of the metric behavior and discuss what it means in the context of deliberative democratic theory. 

\textit{Representation} is a conversation-level measure which measures how the views expressed in a single conversation compare to the breadth of views expressed overall. This measure allows us to assess the overall prevalence and distribution of various oppositional vs. supportive viewpoints across conversations. However, it does not capture whether individual participants directly engage with alternative viewpoints. Nor does it tell us much about the exchange of viewpoints within any given conversation. 
\textit{Fragmentation} in contrast, is a user-level metric which allows us to assess whether and how viewpoints are placed in dialogue with one another, and as such, gives richer meaning and content to the analysis of echo chambers. 
In contrast to previous work, we consider the specific viewpoints user engage with or to which they are exposed to, rather than the worldview or ideology of discussants. We apply these measures to two topics---daylight savings time (DST), which serves as a control, and the more political topic of immigration. We have selected these two topics because they allow us to situate our measures of Representation and Fragmentation within the larger, platform-level context.

On Twitter, which is known for irreverence and overall negativity \cite{tromble2018thanks}, we might expect oppositional claims to dominate on virtually any given topic. Yet, from a democratic perspective, such oppositional stance-taking is not inherently problematic. That people complain a lot about an issue like DST does not spell doom for democracy. Nor would democracy be in jeopardy if those who complain about DST rarely encounter pro-DST views. If, however, we see a more extreme imbalance for a salient political topic such as immigration, then we do have reason for concern.

We find that Twitter conversations contain relatively few viewpoints overall (i.e., most tweets are observational or informational, not stance-taking), when users do express viewpoints on immigration in the U.S., anti-immigration views dominate. This tendency towards oppositional immigration viewpoints is even more extreme than negativity towards DST---suggesting that this is more than a mere reflection of platform culture. Our further findings are troubling:  anti-immigrant views are rarely countered by pro-immigrant views. Where viewpoint interactions do occur, it is largely because pro-immigration views receive anti-immigrant replies. In other words, while pro-immigrant views receive consistent pushback on the platform, anti-immigrant views largely operate within echo chambers. As discussed in greater detail below, these findings are particularly concerning in light of previous research on the role of such echo chambers in generating attitude extremity \cite{baumann2020modeling,benkler2018network,sunstein2018republic}, spirals of silence \cite{chen2018spiral,hampton2014social}, and asymmetric polarization \cite{freelon2020false,kreiss2021social}.

Overall, this work highlights the importance of examining viewpoint diversity, not just ideological diversity. Our measures of Representation and Fragmentation provide tools for examining viewpoint interactions at both the conversational and individual level, and in turn provide important insights into the democratic health of online conversations. By comparing the salient political topic of immigration to the control topic of DST, we not only demonstrate the presence of echo chambers on Twitter, but illustrate how this effect is more extreme for political conversations. \footnote{Code and sample data available at \url{https://github.com/hadarishav/beyond-digital-echo-chambers}}

This work is a collaborative effort of researchers from computational, social, and political backgrounds. With this paper we also want to emphasize the importance of interdisciplinary research to have a well rounded understanding of the problem. This helps in coming up with holistic solutions and interventions that are beyond the capabilities of general machine learning (ML) models which are made from a very computational perspective. Social and political scientists in the team formulated the nuanced labeling of the data, while computer scientists complemented their efforts to build predictive models and derive insights from the data. Only together could we situate the insights in the context of political discourse and what it means for a deliberative democratic society. This highlights the importance of building other tools and applications that can leverage high quality data, instead of just focusing on building yet another ML model.

\section{Related Work}

From a normative perspective, we draw heavily on work in deliberative democracy, which argues that political conversations between citizens form the foundation of democratic life~\cite{mansbridge1999everyday,habermas1984theory,cohen1989deliberation,dryzek2009democratization}. This literature has examined conversation health in various settings \cite{mansbridge2015minimalist,tromble2015are}, but the focus on conversational dynamics has gained renewed attention in light of the forms of interpersonal, group, and mass communication enabled by social media. Measures of the quality of conversations (e.g., toxicity, rationality, and mutual respect) \cite{mansbridge2015minimalist} have perhaps received the most attention, but homophily and echo chambers have also proven important in the literature \cite{lee2014social,aral2009distinguishing}. 

However, most work on online echo chambers focuses on users’ networks \cite{bakshy2015exposure,barbera2015birds,barbera2015tweeting,bastos2018geographic,cinelli2021echo} the content to which users may be exposed---for example, analyzing the political alignment of news outlets based on URLs contained in a post \cite{bakshy2015exposure,garimella2018political}, or the news media accounts users follow \cite{an2014sharing}. Little previous research examines the specific claims or viewpoints that users encounter \cite{schaefer2021argument, cantador2020exploiting}.

While previous studies suggest that people tend to follow user and organizational accounts with similar political leanings \cite{an2014sharing}, post content from ideologically uniform sources \cite{garimella2018political}, and encounter and engage with content from ideologically-aligned news sites \cite{bakshy2015exposure,garimella2018political}, this body of work is unable to tell us whether and to what extent users engage with divergent viewpoints within and across social media conversations.

Recent literature in the news recommender domain has drawn inspiration from the ``Democratic Notions of Diversity'' that focuses on  grand concepts such as democracy, freedom of speech, inclusion, mutual respect and tolerance ~\cite{helberger, Loecherbach, vrijenhoek2021recommenders}. 
In their work, Helberger\cite{helberger} describes four most commonly used theories discussing the democratic role of media: Liberal, Participatory, Deliberative, and Critical model. As mentioned and described earlier we draw heavily on work in deliberative democracy.
Vrijenhoek et. al. \cite{vrijenhoek2021recommenders} propose 5 metrics, adapted from existing Information Retrieval practices to measure viewpoint diversity of ranked lists of recommendations by news recommenders and quantitatively evaluate the various democratic notions of diversity. Inspired from their conceptualization of Representation and Fragmentation we adapt and operationalize the two metrics for social media conversations. We apply the two metrics on Twitter conversations, present an in-depth analysis of the metric behavior and what it means in the context of deliberative democratic theory and echo-chambers. 

This study extends our understanding of political democratic discourse generally, and echo chambers specifically, by studying these viewpoint-based dynamics. One significant computational challenge to viewpoint-based analysis is identifying what viewpoints are present in a conversation. The closest work focuses primarily on natural language inference (NLI) or stance detection (e.g., ~\cite{kuccuk2020stance}). In NLI, for given pairs of sentences (premise and hypothesis), the task is to predict whether the hypothesis given is True, False or not related with respect to the premise. In stance detection a text is labeled as being for, against, or neutral towards some target topic. Some recent work has also focused on determining the strength of stance and the logic of evaluation that reflects the general perspective behind the stance \cite{draws}.
While these approaches are increasingly more accurate in capturing stance, the label inferred by stance detection does not necessarily reflect what we typically mean by ``opinion'' or viewpoint \cite{joseph2021misalignment,sen2020reliability}. In line with democratic discourse theory, we are primarily interested in whether or not a \emph{claim} is made within a text \cite{tromble2015are}.
This is a more subtle notion than stance and implies the presence of an argument, not just an opinion. Here, we therefore develop our own claim detection classifier, as discussed in Section~\ref{sec:classifier}.

While Reply Trees \cite{shugars2019keep, zeng2018microblog, cogan2012reconstruction, nishi2016reply, kumar2010dynamics, choi2015characterizing, glenski2019characterizing} are by far the most common way to model conversation networks, the literature has taken a range of other approaches, such as Mention Graphs \cite{cogan2012reconstruction}, User Graphs \cite{cogan2012reconstruction}, and Conversation Cascade \cite{bagavathi2019examining, bollenbacher2021challenges}. This past work on conversational structure emphasizes the need for both conversation- and individual-level measures. Online conversations take a range of forms, and individuals may have highly divergent experiences based on this overall conversational typology. Therefore, in developing our novel, viewpoint-based approach to examining political discourse, we create two complementary measures of viewpoint diversity---Representation and Fragmentation---which can be meaningfully interpreted across conversational structures.

\section{Data Collection and Classification}

\begin{figure*}[t!]
    \centering
    \includegraphics[scale=0.7]{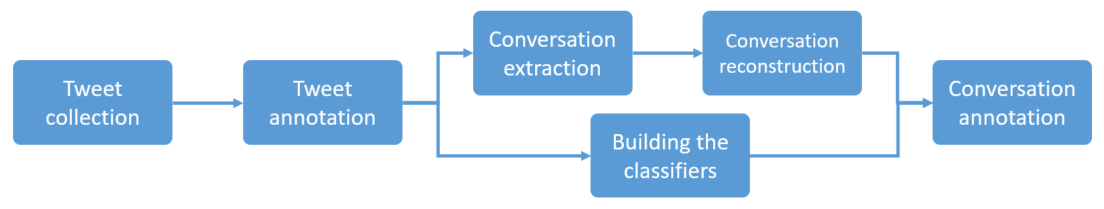}
    \caption{Our pipeline for data collection and classification.}
    \label{fig:paper_flow}
\end{figure*}

As outlined in Figure \ref{fig:paper_flow}, our pipeline for data collection and annotation consists of several steps at both the tweet and conversation level. Each of these steps is described in detail below.

\subsection{Tweet Collection}\label{sec:tweet_collection}
For both topics, daylight savings time (DST) and immigration, we used a keyword-based approach to identify relevant tweets. Keywords were selected by social scientists in a multi-stage process. 
We restricted both samples to English-language tweets and used the Enterprise streaming API
in early 2020 to collect tweets in real time. Due to the time-sensitive nature of DST, the data collection for this topic was conducted between 8$^{th}$ and 10$^{th}$ of March 2020.

\subsection{Tweet Annotation}\label{sec:tweet_annotation}
\textbf{Topical relevance annotation.} 
At this stage in our pipeline, we were interested in retaining only those tweets that were relevant for our topics. 
We therefore developed a relevance classifier for each. These were trained on annotations of 9,814 tweets on DST and 9,931 tweets on immigration respectively. A team of five trained undergraduate annotators from George Washington University manually labeled a random sample of tweets collected based on our keywords. Annotators were told which topic the tweet was collected for and were asked to classify the tweet as either relevant, irrelevant, or ``not English.''

The relevance classifier trained on these annotations is described in further detail in Section~\ref{sec:classifier}.

\textbf{Viewpoint annotation.} Using the relevance classifier, we identify a final seed corpus of 10,529 DST tweets and 15,119 immigration tweets that were then annotated by the same group of undergraduate students for the presence or absence of \textbf{diagnostic claims} and \textbf{counter-claims}. Following previous work \cite{tromble2015are,tromble2016life}, \textit{diagnostic claims} highlight a problem associated with a topic and represent an \textit{oppositional viewpoint} in relation to that topic. For example, a tweet about daylight savings time might contain a diagnostic claim that suggests that DST interrupts sleep schedules. For the \textit{immigration} topic, \textit{diagnostic claims} identified a problem with immigrants or permissive immigration policies. In other words, they were explicitly anti-immigrant/anti-immigration. In contrast to diagnostic claims, \textit{counterclaims} counter the concerns identified in the former. They are considered ``counterclaims'' even when made in isolation from the diagnostic claims they seek to counter. For example, a single tweet that describes how DST helps sleep schedules would be considered to include a counterclaim in our framework (i.e., it logically counters the problem diagnosis that DST interrupts sleep, even if that diagnosis is not explicitly made), as would a tweet that suggests that immigrants benefit the economy (i.e., since it logically counters the problem diagnosis that immigrants harm the economy). 

Pairs of students independently annotated tweets in batches (mean batch size = 302 tweets). 
In order to prevent discrepancies developing across annotators, the student pairs rotated with each batch, with the fifth student in each rotation attending the resolution meetings, observing and sharing any apparent discrepancies with the full team. The team then collectively agreed on a standard and clarified the annotation guidelines accordingly.  
The period in which these standards were being set and updated in the guidelines was treated as a training phase. 

Once the team annotated three batches without any changes to the guidelines \textit{and} inter-annotator agreement was consistently high (above 0.80 for percent agreement and above 0.70 for Krippendorff's alpha), the team began full annotation. The procedures remained the same during this phase, with one student continuing to observe resolution meetings, but no guideline updates were deemed necessary. We measured pairwise inter-annotator agreement based on
whether both annotators agreed that a viewpoint was or was not present. 
For the DST dataset, percent agreement was 0.87 and Krippendorff's alpha was 0.72. 
For the immigration dataset, percent agreement was 0.85 and Krippendorff's alphas was 0.71.

The dataset is still under development i.e. we are adding more fine grained labels. This full dataset will be published in a separate paper with further details about its curation. For this paper, we release a sample of the data for the purposes of reproducibility. We see the main contribution of this work as the operationalization and interpretation of our metrics of viewpoint diversity in the context of social media conversations, described in the subsequent sections.

\subsection{Relevance and Viewpoint Detection Models}\label{sec:classifier}
We made use of recent neural language models~\cite{rogers-etal-2020-primer,nozza2020mask} to build separate classifiers to predict the (1) relevance of the tweets to our topics and (2) to determine whether the tweets contain any diagnostic claims, counterclaims, or no viewpoint. We built these classifiers separately for each topic, resulting in a total of four trained classifiers.

We used BERTweet~\cite{bertweet} a large language model pre-trained on tweets and fine-tuned it with our datasets  (BERTweet is used as encoder to which we add an additional classification layer). For each topic and classification task, we trained a model for 4 epochs on 80\% of the data, with a batch size of 32. We used 10\% of the data to evaluate it every 20 steps. Finally, we picked the model with the lowest validation loss and we evaluate it on the last 10\% of the data.

Overall, our models return relatively accurate results across tasks and topics, including the hardest classification tasks. The \textbf{relevance model} for \textit{DST} had a macro-F1 score of .95 , with a precision of .92 and a recall of .98. Similarly, the trained relevance model for \textit{immigration} had a macro-F1 score of .92, identifying relevant tweets with a precision of .92 and a recall of .92. \textbf{Viewpoint classification} proved to be a harder task, but our models achieved a macro-F1 of .80 for \textit{DST} and a macro-F1 of .80 for \textit{immigration}. For both models, identification of counterclaims was particularly challenging, with this class achieving a precision of .75 and a recall of .78 for DST, and a precision of .70 and a recall of .74 for immigration. 

\subsection{Conversation extraction and reconstruction}\label{sec:conversation_extraction}
\begin{figure}[t!]
    \centering
    \includegraphics[scale=0.15]{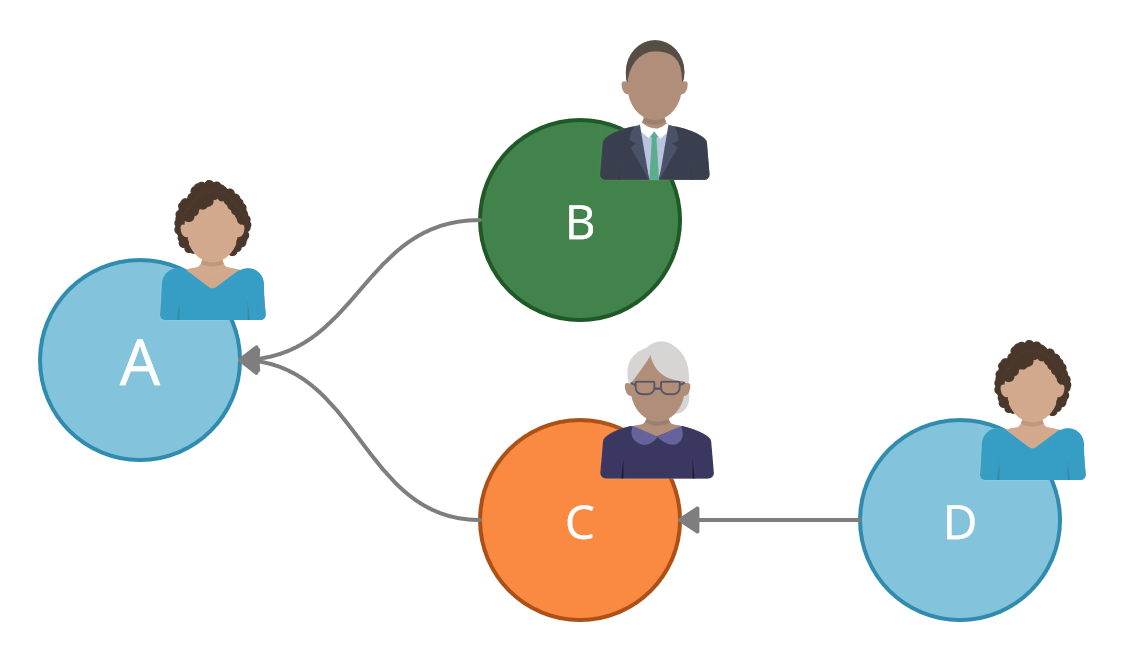}
    \caption{A conversation network containing four tweets of A, B and C, and D. The initial tweet A is the root node of the tree. Tweets B and C are both replies to A, while D is a reply to tweet C. Tweets A and D are written by the same author.} 
    \label{fig:conversation_network}
\end{figure}

\textbf{Representing Conversations.} After the annotation of our seed tweets, we collected and reconstructed the conversations of which the seed tweets are a part. To do this, we first operationalized the notion of a  ``conversation''. Of the four approaches summarized in the related work, we follow the model of reply trees. \footnote{We consider only replies. We do not consider other forms of engagement like quote tweet and retweet.} This choice is suitable for examining the degree to which conflicting views come into contact with each other, since they show exactly what content is shared \emph{in response} to other content. 
More formally, reply trees are directed, acyclic graphs with at least two nodes. Each node represents a single tweet, and each edge shows a reply from a newer tweet to an older tweet (c.f., Figure~\ref{fig:conversation_network}). 

Any given tweet may receive any number of replies, but may only be made in response to at most one tweet. Hence the resulting graph is strictly acyclic and contains no loops. The root node $A$ in Figure~\ref{fig:conversation_network} has an out-degree of zero, indicating that it is not in response to any other tweet. All other tweets have an out-degree of 1, indicating that they have exactly one parent to which they are responding. Furthermore, the nodes can have any in-degree value--e.g., can receive any number of replies. A tweet which receives no replies (in-degree of zero), is defined as a leaf. Note also that each tweet is associated with a user, and each user may author any number of tweets.

\textbf{Conversation Extraction.}
We used the Academic Research Track of the Twitter API to reconstruct conversations. Under this new track of the API, each tweet is associated with a \textit{conversation\_id} which is shared across all tweets in a unique reply tree. This approach allows us to make calls for each conversation. 
 
Based on initial API calls using both DST and immigration seed tweets, we decided to retrieve a maximum of 50 tweets per \textit{conversation\_id}. This limit allows us to capture the majority of tweets connected to each conversation while keeping the computational task tractable. We further defined a conversation to require at least two different tweets from at least two different authors. For the purposes of this analysis we discard any singleton tweets which received no replies, as well as any single-author threads.

\textbf{Conversation Reconstruction.} After extracting the conversation tweets, we reconstructed the conversation tree by using the \textit{referenced\_tweets} information from each retrieved tweet object. For each tweet, we examined whether it had a \textit{referenced\_tweets} of type \textit{replied\_to}

If this value is None, the given tweet is a root node and has out-degree of zero. Otherwise, this field lists the unique \textit{id} of that tweet's parent---the single tweet to which the examined tweet is replying. We then reconstructed each reply tree by iteratively attaching each child to its parent. For example, we connect tweet A to tweet B, if $\textit{referenced\_tweets} \rightarrow \textit{id}$ field in the former is equal to the \textit{id} field in the latter. \footnote{Each retrieved tweet object includes information on both the retrieved tweet itself and its parent, if there is one. This means that for some of our conversations, the final size of the retrieved conversation was larger than the size limit of 50 that we set.}

\textbf{Conversation dataset.} Our seed corpus of 10,531 DST tweets is ultimately associated with 1,756 unique conversations.
On average, conversations about immigration tended to be longer than those about DST. Our 9,667 seed immigration tweets were associated with 404 unique conversations. \footnote{$\approx 3\%$ and $\approx 14\%$ of the DST and immigration conversations respectively were greater than our set threshold of 50 tweets. We retain 50 tweets each for such conversations.}

\subsection{Conversation annotation}\label{sec:conversation_annotation}

Given reconstructed conversations, the final step was to identify the relevance and viewpoint status of each tweet in our conversational corpus.  Table \ref{tab:conversations_basic_information} provides a summary of the structural features of our final corpus of conversations, along with the distribution of annotation labels for each topical dataset. 
The first step for obtaining the labels was to pass every tweet through its topic's relevance model, which labeled it as either relevant, not relevant or not English (Section~\ref{sec:classifier}).

Next, we passed tweets through our viewpoint detection models that marked tweets as containing a diagnostic claim, a counterclaim, or none. The relevance and claim detection models are described in Section~\ref{sec:classifier}.
We then aggregated our classification output into four distinct labels.
Tweets that are ``irrelevant" (L1). Tweets that are relevant but have no diagnostic claim or counterclaim are labeled as ``no viewpoint" (L2). Tweets that are identified as including a diagnostic claim are labeled as ``Diagnostic claim" (L3) regardless of the output of our relevancy classifier. \footnote{Certain tweets might be irrelevant but still contain a claim in the context of a conversation.} Similarly, all tweets that are identified as containing a ``counterclaim'' are labeled as (L4). 

\begin{table*}[t!]
\centering
\resizebox{\textwidth}{!}{%
\begin{tabular}{@{}lcccccccc@{}}
\toprule
\textbf{Topic} & \multicolumn{4}{c}{\textbf{Structural Information}} & \multicolumn{4}{c}{\textbf{Annotations}} \\ 
 &
  \multicolumn{1}{l}{\# of conversations} &
  \multicolumn{1}{l}{\# of nodes} &
  \multicolumn{1}{l}{\# of edges} &
  \multicolumn{1}{l}{\# of distinct users} &
  \begin{tabular}[c]{@{}c@{}}Irrelevant\\ (L1)\end{tabular} &
  \begin{tabular}[c]{@{}c@{}}No viewpoint\\ (L2)\end{tabular} &
  \multicolumn{1}{l}{\begin{tabular}[c]{@{}l@{}}Diagnostic\\Claim (L3)\end{tabular}} &
  \begin{tabular}[c]{@{}c@{}}Counter Claim\\ (L4)\end{tabular} \\ \midrule
DST            & 1756       & 15362       & 13606       & 10578      & 86.85\%    & 6\%   & 4.04\%  & 3.1\%   \\
Immigration    & 404        & 13304       & 12900       & 8611       & 78.43\%   & 9.86\%   & 7.7\%  & 4.01\%  \\ \bottomrule
\end{tabular}%
}
\caption{Structural information of conversations and overall distribution of labels in both topics}
\label{tab:conversations_basic_information}
\end{table*}

\section{Viewpoint Diversity Measures}\label{sec:measures}
To better evaluate viewpoint diversity, we introduce a conversation-level measure of Representation and a user-level measure of Fragmentation. Each measure provides insight into the degree to which conflicting views come into contact with each other, either within a conversation or for individual participants. In Section~\ref{sec:results}, we  present the results of these metrics for our DST and immigration datasets.

\subsection{Fragmentation Diversity}
\paragraph{Definition.} We define the user-level metric of Fragmentation as the complement of the overlap between users’ viewpoint \cite{vrijenhoek2021recommenders}. This metric aims to capture the extent to which individuals within a conversation are exposed to different viewpoints. A value of 1 means people are exposed to maximally different viewpoints, while 0 means people are exposed to the same viewpoints.

\begin{figure}[t!]
    \centering
    \subfigure[]{\includegraphics[width=0.30\textwidth]{
    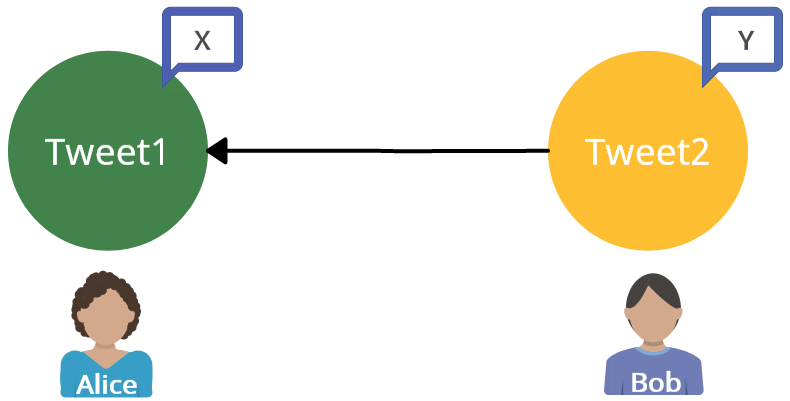}
    \label{fig:viewpoint_exposure}
    } 
    \subfigure[]{\includegraphics[width=0.30\textwidth]{
    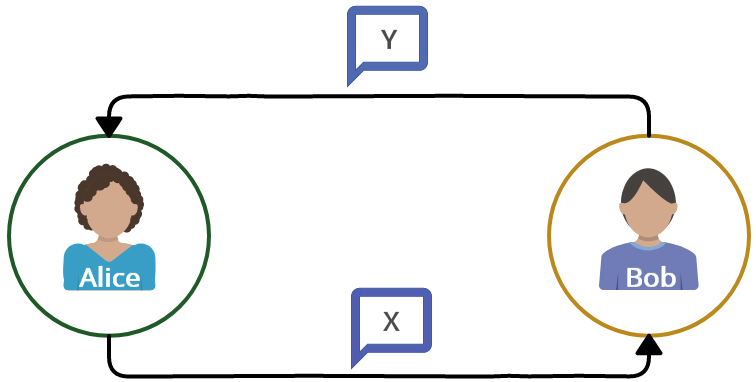}
    \label{fig:viewpoint_network}
    }
    \subfigure[]{\includegraphics[width=0.17\textwidth]{
    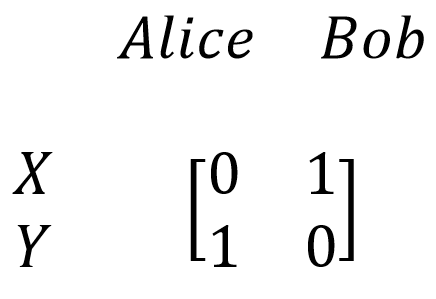}
    \label{fig:viewpoint_matrix}
    } 
    \subfigure[]{\includegraphics[width=0.13\textwidth]{
    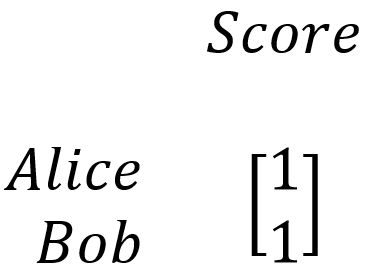}
    \label{fig:scorer-example}
    } 
    \caption{Measuring the Fragmentation score in an exemplar conversation: (a) The exposure of different viewpoints in a conversation. In this conversation Bob is exposed to the viewpoint X and Alice is exposed to the viewpoint Y, (b) The corresponding viewpoint network to the conversation between Alice and Bob, (c) The viewpoint matrix corresponding to the viewpoint network, and (d) the Fragmentation score for Alice and Bob.}
    \label{fig:Fragmentation_score_exemplar}
\end{figure}

\paragraph{Operationalization.} We define exposure at the level of pair-wise (dyadic) interactions, considering a user to be exposed to a viewpoint if they are replying to a certain viewpoint (e.g., X), or if they receive a reply with a certain viewpoint (e.g., Y). \footnote{We do not consider the user's own viewpoint since we want to observe what other viewpoints a user is exposed to or engages with in a conversation.}  Figure \ref{fig:viewpoint_exposure} illustrates an imaginary conversation in which Alice posts Tweet1 containing viewpoint X and then Bob replies with Tweet2 containing viewpoint Y. In this scenario, Bob is exposed to Alice's viewpoint through Tweet1 and, after he posts Tweet2, Alice is exposed to Y.  If a third user responded to Bob with viewpoint Z, Bob would be exposed to Z, but Alice would not.

To measure the degree of overlap in viewpoint exposure, we transform our conversation network into a \textit{viewpoint network}.  A viewpoint network is a multi-directed graph in which each node represents a unique user and each edge shows who is exposed to whom and with what viewpoint. In this network, in-degree represents viewpoints a user is exposed to, while out-degree indicates the views the user is disseminating. Figure \ref{fig:viewpoint_network} demonstrates the corresponding viewpoint network to the conversation network between Alice and Bob.

We next construct the \textit{matrix representation} of this viewpoint network, as shown in Figure \ref{fig:viewpoint_matrix}. Every column in this viewpoint matrix represents a single user, while each row indicates a single viewpoint. Here, we use the viewpoint labels of \textbf{L1} (irrelevant), \textbf{L2} (no viewpoint), \textbf{L3} (diagnostic claim), and \textbf{L4} (counterclaim) consistently across all conversations. A conversation with U unique authors will therefore have a viewpoint matrix of size 4 x U. This is a positive, weighted matrix in which a user may be exposed to a viewpoint any number of times. Each of these column vectors describe a single user's position in a shared viewpoint space. Similarity between vectors then reflects similarity between the viewpoints that the corresponding users are exposed to.
Therefore, to compute the Fragmentation score, we first calculate the similarity between every pair of user vectors (i.e., columns) in the viewpoint matrix.

We calculate this using cosine similarity which ranges between 0 (no overlap) to 1 (complete overlap). For each user in a given conversation, this results in a list of U-1 similarity scores. Next, we take the mean of each user's pairwise similarity scores, and finally subtract this value from 1 since Fragmentation and overlap (similarity) have an inverse relationship. Recall, a Fragmentation value closer to 1 means that the user is exposed to maximally different viewpoints from their conversational peers, and Fragmentation value closer to 0 means that the user is exposed to maximally similar viewpoints from their conversational peers.

\subsection{Representation Diversity}
\paragraph{Definition.} For our Representation metric, we adapt the definition and operationalization from \cite{vrijenhoek2021recommenders} for the context of social media conversations. Representation compares the views expressed in a single conversation to the breadth of views expressed for the topic overall. Representation thereby provides a \textit{conversation-level metric} which denotes the degree to which conversations are restricted to certain views or capture the diversity of possible opinions. A Representation score of 0 indicates that the distribution of viewpoints in a conversation is similar to the overall distribution of viewpoints in the topical data pool. As we move towards a Representation score of 1 the discrepancies between a given conversation and the data pool increases, while scores closer to 0 indicate that the conversation is ``typical'' or representative in terms of viewpoint diversity (in the context of a given topic).

\paragraph{Operationalization.} We compute Representation by measuring the Kullback-Leibler (KL) divergence between the probability distribution of the viewpoint categories in a single conversation to the viewpoint distribution in the entire pool of conversations data for a given topic. Here, the possible viewpoint categories again refers to the labeling system described in Section~\ref{sec:tweet_annotation}. 
We measure the KL divergence between the two distributions and then normalize the value for each conversation with the maximum KL divergence value obtained. 

\section{Results and Analysis}\label{sec:results}

In this section, we report our results on the Fragmentation and Representation metrics for Twitter conversations for two topics: immigration and daylight savings.

\subsection{Fragmentation Diversity}
\begin{figure}[t!]
    \centering
    \subfigure[]{\includegraphics[scale=0.3]{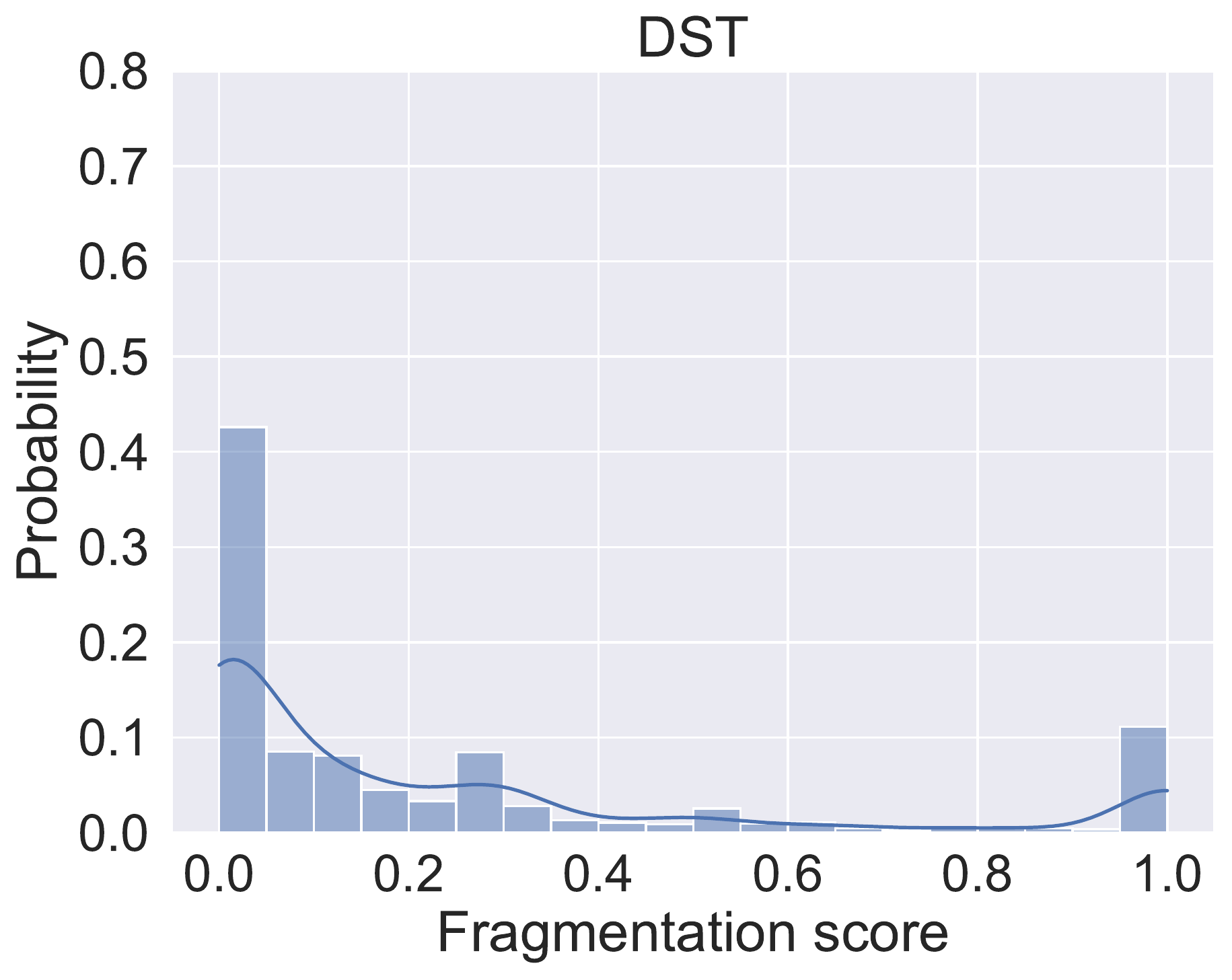}}
    \subfigure[]{\includegraphics[scale=0.3]{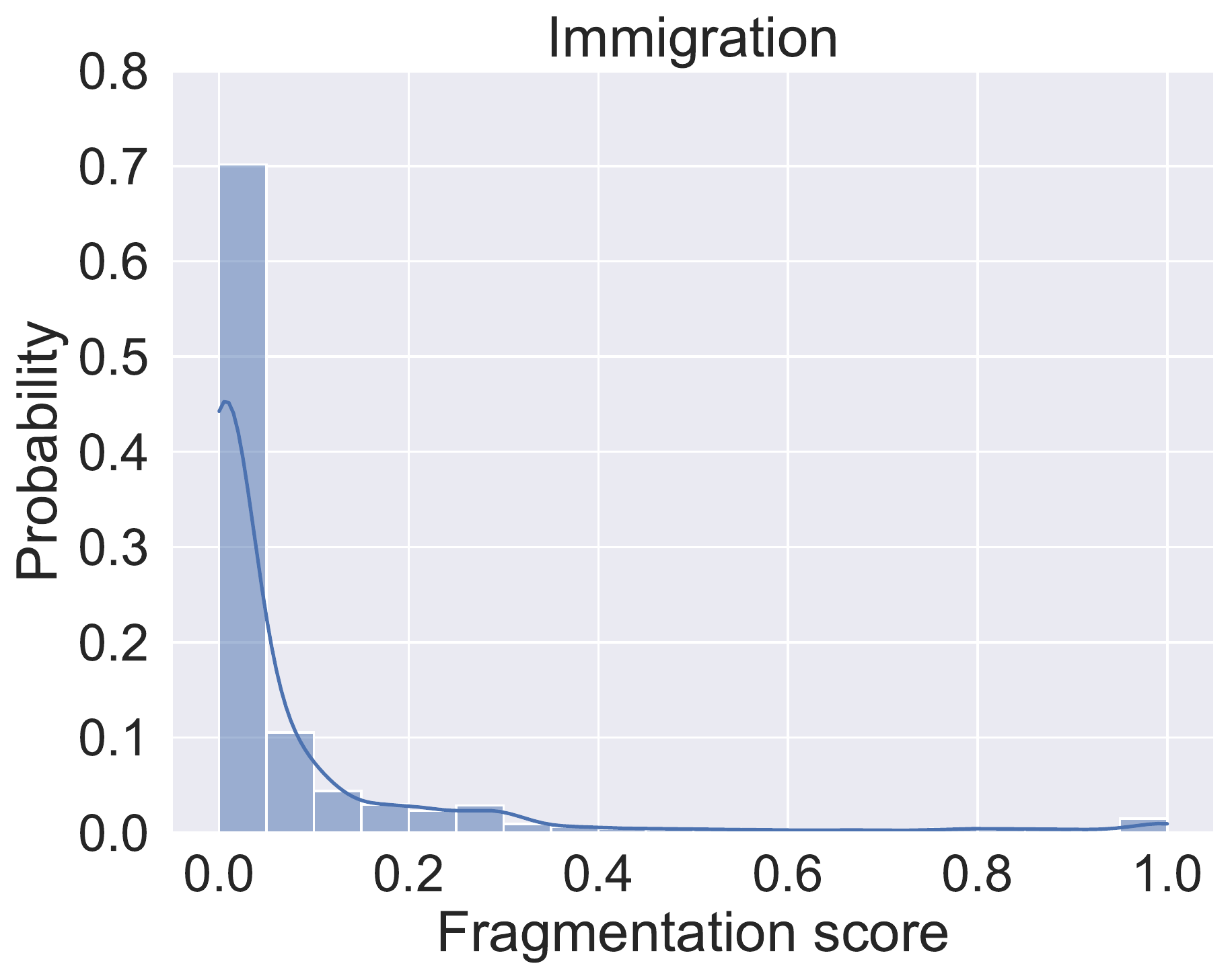}} 
    \caption{The distribution of Fragmentation for daylight savings time (DST) (a) and immigration (b) conversations. The x-axis shows a Fragmentation score per user. (\textit{Fragmentation is defined as the complement of the overlap between users’ viewpoint. A high fragmentation value means people are exposed to maximally different viewpoints, while a low value means people are exposed to the same viewpoints in a conversation.})}
    \label{fig:Fragmentation}
\end{figure}

Figure \ref{fig:Fragmentation} shows the distribution of Fragmentation values for both topics. As described in Section~\ref{sec:measures}, Fragmentation is the complement of overlap between individuals' viewpoint exposure in a conversation. 
Recall that a user with a Fragmentation score closer to 0 is exposed to the same viewpoints as their peers.\footnote{Note that that even if a user is exposed to exactly one viewpoint, their Fragmentation score could be 0, if their peers are exposed to exactly one viewpoint as well (a limitation addressed by the metric of Representation).} In DST conversations, more than 40\% of users have Fragmentation scores between 0\textemdash0.05, indicating that users in these conversations have a high overlap among the exposed viewpoints. In contrast, a user with a Fragmentation score closer to 1 is exposed to viewpoints different from their conversational peers.  On the other hand, we also see that over 10\% of users in these conversations have a Fragmentation score of over 0.95, indicating that a notable number of users do often diverge from the majority viewpoints being discussed in the conversations.
Interestingly, this finding is even more extreme within the conversations about immigration. Nearly 70\% of users have a Fragmentation score between 0\textemdash0.05, and virtually none have a Fragmentation score near 1.

It is not evident whether the prevalence of L1 (``Irrelevant'') viewpoints in our dataset should be considered noise. Therefore, we also compute our results of Fragmentation without considering L1 viewpoints. That is, we did not use the L1 values from the viewpoint matrix when calculating similarity between users. We notice a similar distribution as before for Fragmentation values for both topics without L1 viewpoints.

\subsection{Representation Diversity} 
\begin{figure}[t!]
    \centering
    \subfigure[]{\includegraphics[scale=0.3]{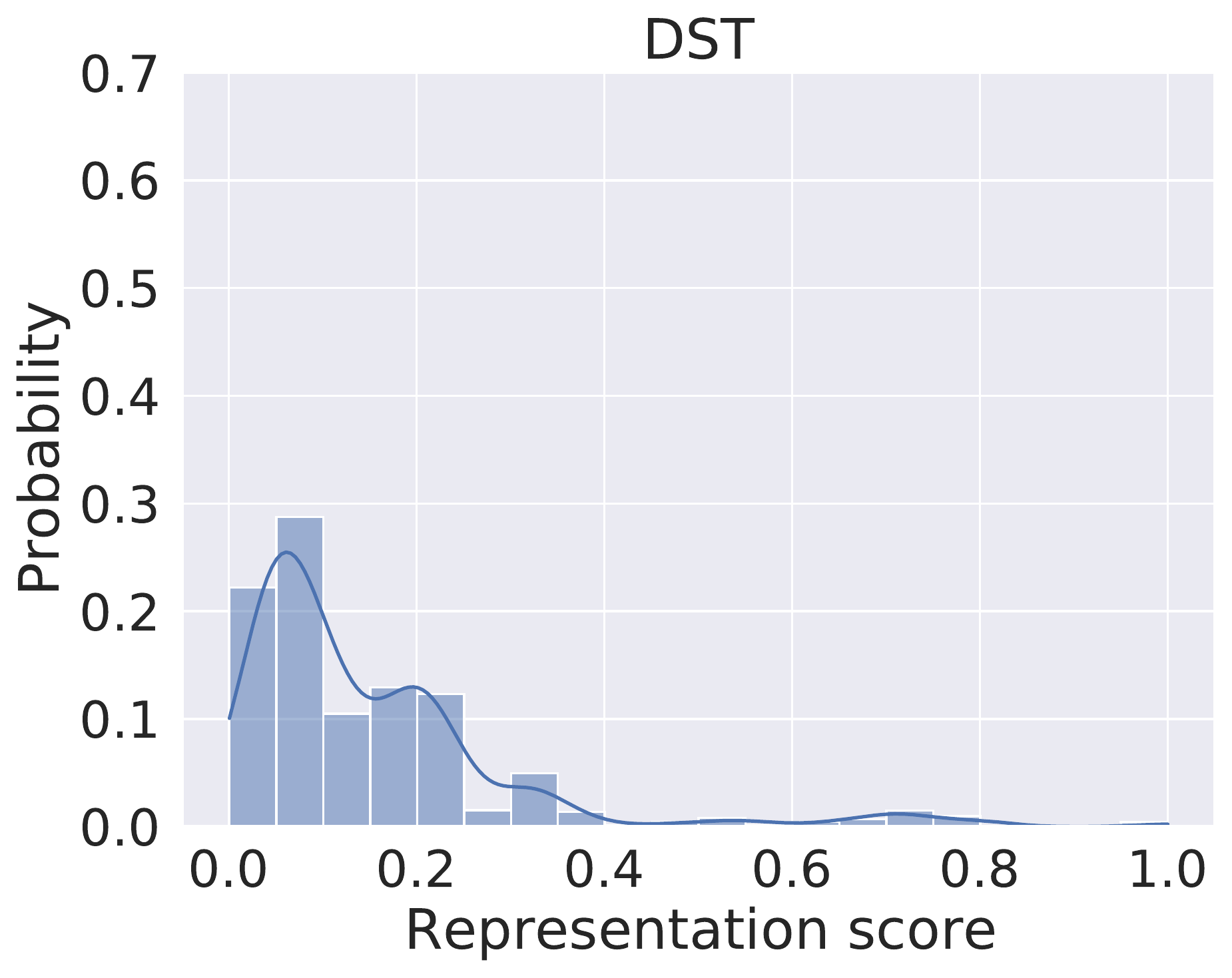}} 
    \subfigure[]{\includegraphics[scale=0.3]{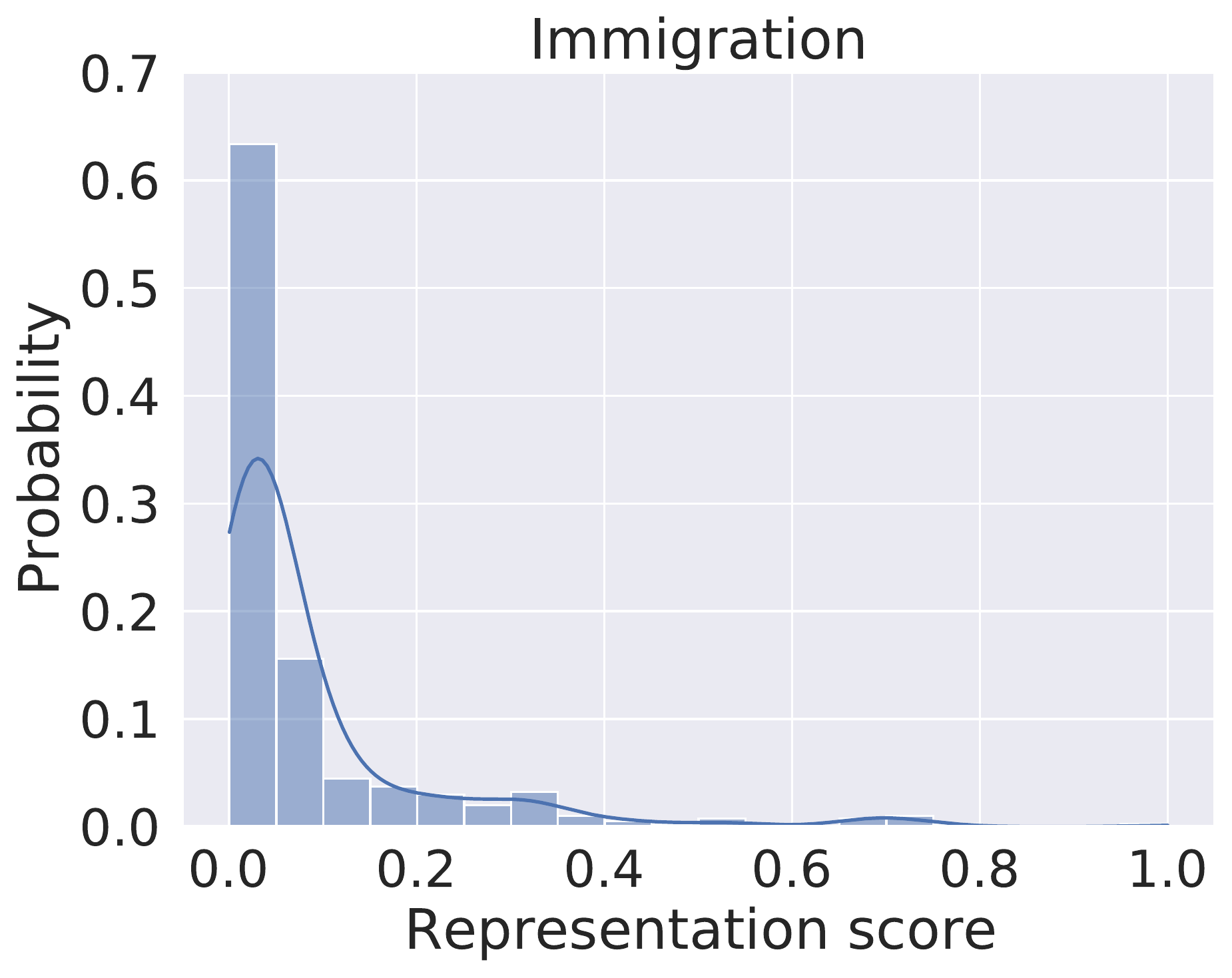}} 
    \caption{The distribution of Representation for daylight savings time (DST) (a) and immigration (b) conversations. The x-axis shows a Representation score per conversation. (\textit{Representation compares the views expressed in a single conversation to the breadth of views expressed for the topic overall. As we move from lower to higher value the divergence between the distribution of viewpoints in the conversation and the topical data pool increases.})}
    \label{fig:Representation}
\end{figure}

Representation is a conversation-level diversity metric
that compares the views expressed in a single conversation to the breadth of views expressed for the topic overall. Figure \ref{fig:Representation} shows the distribution of Representation scores for both topics.  Recall that a score close to 0 indicates that a conversation's distribution of viewpoints matches the distribution in the overall pool of viewpoints for the topic. 
We see that over 20\% of DST conversations have a low level of diversity (score between 0\textemdash0.05), whereas over 60\% of conversations about immigration fall into that bin.
This suggests that the majority of individual immigration conversations mirror the viewpoint distribution compared to all conversations on immigration (described in Table~\ref{tab:conversations_basic_information}). In contrast, individual conversations about DST tend to have more variability from the full set of conversations on DST.
However, it is also interesting to note that Representation scores for immigration also have a much longer tail---indicating that while these conversations are most likely to to be similar to the pool distribution, some of the deviations are also extreme. 

Again, we consider the influence of the L1 viewpoint (irrelevant) on the metric performance. Similar to Fragmentation, we compute Representation results without considering L1 viewpoints. Operationally this means that we did not consider the L1 values when computing the KL divergence between the distribution of the pool and the conversation. We again observed similar distributions as before for Representation values for both topics without L1 viewpoints. 

\subsection{Dyadic Interactions}
To investigate how the stance-taking viewpoints i.e., oppositional claim (L3) and supporting counterclaim (L4) engage with each other in a conversation we further examine their pair-wise (dyadic) interactions. To focus on substantive conversation, we ignore all other dyadic interactions (with L1 and L2).  
We calculate the conditional probability of $P(L_i|L_j)$ where $L_i$ and $L_j$ represents labels with $i, j \in \{3,4\}$. 
It represents the likelihood of a reply with label $L_i$ to a tweet with label $L_j$. 
For instance, $P(L_4|L_3)=0.33$ means that, given a tweet with an oppositional claim, the likelihood that a replying tweet has a supporting counterclaim is $0.33$.  

Here, we see evidence for echo-chambers occurring, particularly among users expressing the oppositional (L3) viewpoints for immigration. For the control topic of DST, we see that when beginning with tweets that contain the oppositional viewpoint (L3)---a responding tweet has a 62\% chance of similarly containing the oppositional viewpoint. For immigration, that probability rises to 77\%. In other words, more than three-quarters of responses to anti-immigrant claims also include anti-immigrant viewpoints. Those who voice support for immigrants (L4) in contrast have a lower likelihood of receiving a L4 reply (49\%). In other words, we can see this as more echo-chamber interactions on the oppositional side for immigration. 

What is perhaps more worrying is that those who voice support for immigrants (L4) also have a 51\% chance of receiving an anti-immigrant reply (L3). Again, we see that this effect is even more extreme for our political issue than it is for DST. A user who supports DST (L4) has only a 38\% chance of receiving an anti-DST (L3) reply. 

Compared to conversations about DST, we see that conversations espousing anti-immigrant viewpoints tend to take place in echo chambers, while comments supporting immigrants are frequently met by anti-immigrant retorts. While the difference in response may not be dramatic across the two topics, it is concerning that opposition to immigration appears to go largely unchallenged, while support for immigration receives regular pushback. 

\section{Conclusion}
In this section, we discuss the broader implications of our findings, limitations of our data and methods, possible directions for future work, and some additional applications of the measures developed.

\paragraph{Broader Implications.} Bringing these findings together, our work suggests that oppositional  viewpoints, such as those opposing daylight savings or immigrants and immigration, are more common in Twitter conversations than supportive viewpoints and are more likely to exist within echo chambers. While this can be partially attributed to the overall culture of the platform under study, we see that this effect is even more pronounced for immigration than for the less politically salient topic of DST. Our conversation-level measure of Representation shows that conversations about both immigration and DST reflect the distribution of opinions in the full population. 

Our more fine-grained, individual-level measure of Fragmentation further shows that individual users tend to have very little variability in the viewpoints to which they are exposed. This effect is more pronounced for immigration than for DST. Only when looking at these metrics \underline{together} can we start to diagnose the presence of echo chambers: a low Representation value of most conversations show that they have skewed distribution and consist of the majority (irrelevant or oppositional) viewpoints. While, a low Fragmentation score for most users show that they are exposed primarily to these majority viewpoints.

Our analysis of pair-wise responses also indicates that for the topic of immigration users who share oppositional claims are most likely to receive like-minded replies, while users who make counterclaims receive both like-minded and oppositional replies. This suggests distinctly divergent user experiences---users who make diagnostic claims,  may interact primarily in echo chambers while users who make counterclaims in support of the issue at hand may be regularly forced to defend and justify their views. It is particularly troubling that this dynamic is more  prevalent within our politically salient topic of immigration (compared to the topic of DST)---suggesting that this pattern may be more pronounced for political talk.

Though this study alone cannot diagnose the extent to which such echo chambers harm democratic discourse on Twitter, previous research suggests that online echo chambers can promote attitude extremity \cite{baumann2020modeling,benkler2018network,sunstein2018republic}. And the one-sided nature of the echo chambers we find in this study points to further concerns about the silencing of pro-immigrant voices, as fear of (or exhaustion from) backlash may lead to a spiral of silence \cite{chen2018spiral,hampton2014social}. And it bolsters larger concerns that \textit{asymmetric} polarization offers a breeding ground for intolerance of and discrimination against marginalized communities \cite{freelon2020false,kreiss2021social}.

\paragraph{Limitations and Future Work.}
Rate limits of the Twitter API\footnote{Twitter Developer Platform. 2022. Academic Research Track. (2022).  \url{https://developer.twitter.com/en/products/twitter-api/academic-research}. Accessed: 2022-01-19.} continue to be a major bottleneck for conversational research.
This challenge is compounded by deleted tweets or accounts marked as private. 
This missing data can then lead to disconnected components in conversations, which is a notable limitation for conversational research. For the datasets in this paper, we found that 18\% of DST while 63\% of immigration conversations resulted in disconnected components.

Firstly, we chose a specific representation of viewpoint labels. In future work, we will expand our datasets by incorporating even more fine-grained labels that allow us to explore the range and diversity of viewpoints \textit{within} diagnostic and counterclaim categories.
Secondly, even though our analysis with and without irrelevant viewpoint shows similar patterns, the prevalence of irrelevant tweets in conversations is a considerable limitation when analysing social media conversations. Thirdly, we rely on ML models for our analysis. As is the case when using any automated system, some predictions might be incorrect, and there might be unwanted biases learned by the system.
Lastly, the robustness of our results could be further investigated by exploring other design choices and operationalization metrics. There is an imbalance in the number of conversations for DST and immigration. More immigration conversations can be extracted for a more robust comparison. Another direction could be to study other ways of defining a conversation network and how that impacts the viewpoint measures. 
Future work might also supplement Fragmentation and Representation with other dimensions of viewpoint diversity.

\paragraph{Additional Applications}
 
Recent studies have highlighted the importance of carefully curating and documenting datasets on which language models are trained \cite{bender-friedman-2018-data, timnit, parrots}. Bender et al. \cite{parrots} argue that data dumps taken from the Internet retain only hegemonic viewpoints overrepresenting younger users and those
from developed countries. The authors propose datasets should be created with a thoughtful process such that they are diverse in terms of the viewpoints represented. Our measures of Representation and Fragmentation can act as essential dimensions of viewpoint diversity for evaluating conversational datasets. Furthermore, in capturing the diverging viewpoint dynamics of different types of conversations, these measures could potentially be used to help identify particularly contentious topics. Detecting low viewpoint diversity could be especially valuable for identifying both individual accounts and semi-organized efforts to intentionally and regularly provide oppositional replies without ever engaging in good-faith exchange.

\section*{Acknowledgments} 

This research was supported by a gift from Twitter, Inc.  Twitter exerted no control over the direction or findings of the research.

\bibliographystyle{ACM-Reference-Format}
\bibliography{acmart.bib}


\begin{thebibliography}{49}


\ifx \showCODEN    \undefined \def \showCODEN     #1{\unskip}     \fi
\ifx \showDOI      \undefined \def \showDOI       #1{#1}\fi
\ifx \showISBNx    \undefined \def \showISBNx     #1{\unskip}     \fi
\ifx \showISBNxiii \undefined \def \showISBNxiii  #1{\unskip}     \fi
\ifx \showISSN     \undefined \def \showISSN      #1{\unskip}     \fi
\ifx \showLCCN     \undefined \def \showLCCN      #1{\unskip}     \fi
\ifx \shownote     \undefined \def \shownote      #1{#1}          \fi
\ifx \showarticletitle \undefined \def \showarticletitle #1{#1}   \fi
\ifx \showURL      \undefined \def \showURL       {\relax}        \fi
\providecommand\bibfield[2]{#2}
\providecommand\bibinfo[2]{#2}
\providecommand\natexlab[1]{#1}
\providecommand\showeprint[2][]{arXiv:#2}

\bibitem[An et~al\mbox{.}(2014)]%
        {an2014sharing}
\bibfield{author}{\bibinfo{person}{Jisun An}, \bibinfo{person}{Daniele
  Quercia}, \bibinfo{person}{Meeyoung Cha}, \bibinfo{person}{Krishna Gummadi},
  {and} \bibinfo{person}{Jon Crowcroft}.} \bibinfo{year}{2014}\natexlab{}.
\newblock \showarticletitle{Sharing political news: the balancing act of
  intimacy and socialization in selective exposure}.
\newblock \bibinfo{journal}{\emph{EPJ Data Science}}  \bibinfo{volume}{3}
  (\bibinfo{year}{2014}), \bibinfo{pages}{1--21}.
\newblock


\bibitem[Aral et~al\mbox{.}(2009)]%
        {aral2009distinguishing}
\bibfield{author}{\bibinfo{person}{Sinan Aral}, \bibinfo{person}{Lev Muchnik},
  {and} \bibinfo{person}{Arun Sundararajan}.} \bibinfo{year}{2009}\natexlab{}.
\newblock \showarticletitle{Distinguishing influence-based contagion from
  homophily-driven diffusion in dynamic networks}.
\newblock \bibinfo{journal}{\emph{Proceedings of the National Academy of
  Sciences}} \bibinfo{volume}{106}, \bibinfo{number}{51}
  (\bibinfo{year}{2009}), \bibinfo{pages}{21544--21549}.
\newblock


\bibitem[Bagavathi et~al\mbox{.}(2019)]%
        {bagavathi2019examining}
\bibfield{author}{\bibinfo{person}{Arunkumar Bagavathi},
  \bibinfo{person}{Pedram Bashiri}, \bibinfo{person}{Shannon Reid},
  \bibinfo{person}{Matthew Phillips}, {and} \bibinfo{person}{Siddharth
  Krishnan}.} \bibinfo{year}{2019}\natexlab{}.
\newblock \showarticletitle{Examining untempered social media: analyzing
  cascades of polarized conversations}. In
  \bibinfo{booktitle}{\emph{Proceedings of the 2019 IEEE/ACM International
  Conference on Advances in Social Networks Analysis and Mining}}.
  \bibinfo{pages}{625--632}.
\newblock


\bibitem[Bakshy et~al\mbox{.}(2015)]%
        {bakshy2015exposure}
\bibfield{author}{\bibinfo{person}{Eytan Bakshy}, \bibinfo{person}{Solomon
  Messing}, {and} \bibinfo{person}{Lada~A Adamic}.}
  \bibinfo{year}{2015}\natexlab{}.
\newblock \showarticletitle{Exposure to ideologically diverse news and opinion
  on Facebook}.
\newblock \bibinfo{journal}{\emph{Science}} \bibinfo{volume}{348},
  \bibinfo{number}{6239} (\bibinfo{year}{2015}), \bibinfo{pages}{1130--1132}.
\newblock


\bibitem[Barber{\'a}(2015)]%
        {barbera2015birds}
\bibfield{author}{\bibinfo{person}{Pablo Barber{\'a}}.}
  \bibinfo{year}{2015}\natexlab{}.
\newblock \showarticletitle{Birds of the same feather tweet together: Bayesian
  ideal point estimation using Twitter data}.
\newblock \bibinfo{journal}{\emph{Political analysis}} \bibinfo{volume}{23},
  \bibinfo{number}{1} (\bibinfo{year}{2015}), \bibinfo{pages}{76--91}.
\newblock


\bibitem[Barber{\'a} et~al\mbox{.}(2015)]%
        {barbera2015tweeting}
\bibfield{author}{\bibinfo{person}{Pablo Barber{\'a}}, \bibinfo{person}{John~T
  Jost}, \bibinfo{person}{Jonathan Nagler}, \bibinfo{person}{Joshua~A Tucker},
  {and} \bibinfo{person}{Richard Bonneau}.} \bibinfo{year}{2015}\natexlab{}.
\newblock \showarticletitle{Tweeting from left to right: Is online political
  communication more than an echo chamber?}
\newblock \bibinfo{journal}{\emph{Psychological science}} \bibinfo{volume}{26},
  \bibinfo{number}{10} (\bibinfo{year}{2015}), \bibinfo{pages}{1531--1542}.
\newblock


\bibitem[Bastos et~al\mbox{.}(2018)]%
        {bastos2018geographic}
\bibfield{author}{\bibinfo{person}{Marco Bastos}, \bibinfo{person}{Dan Mercea},
  {and} \bibinfo{person}{Andrea Baronchelli}.} \bibinfo{year}{2018}\natexlab{}.
\newblock \showarticletitle{The geographic embedding of online echo chambers:
  Evidence from the Brexit campaign}.
\newblock \bibinfo{journal}{\emph{PloS one}} \bibinfo{volume}{13},
  \bibinfo{number}{11} (\bibinfo{year}{2018}), \bibinfo{pages}{e0206841}.
\newblock


\bibitem[Baumann et~al\mbox{.}(2020)]%
        {baumann2020modeling}
\bibfield{author}{\bibinfo{person}{Fabian Baumann}, \bibinfo{person}{Philipp
  Lorenz-Spreen}, \bibinfo{person}{Igor~M Sokolov}, {and}
  \bibinfo{person}{Michele Starnini}.} \bibinfo{year}{2020}\natexlab{}.
\newblock \showarticletitle{Modeling echo chambers and polarization dynamics in
  social networks}.
\newblock \bibinfo{journal}{\emph{Physical Review Letters}}
  \bibinfo{volume}{124}, \bibinfo{number}{4} (\bibinfo{year}{2020}),
  \bibinfo{pages}{048301}.
\newblock


\bibitem[Bender and Friedman(2018)]%
        {bender-friedman-2018-data}
\bibfield{author}{\bibinfo{person}{Emily~M. Bender} {and}
  \bibinfo{person}{Batya Friedman}.} \bibinfo{year}{2018}\natexlab{}.
\newblock \showarticletitle{Data Statements for Natural Language Processing:
  Toward Mitigating System Bias and Enabling Better Science}.
\newblock \bibinfo{journal}{\emph{Transactions of the Association for
  Computational Linguistics}}  \bibinfo{volume}{6} (\bibinfo{year}{2018}).
\newblock


\bibitem[Bender et~al\mbox{.}(2021)]%
        {parrots}
\bibfield{author}{\bibinfo{person}{Emily~M. Bender}, \bibinfo{person}{Timnit
  Gebru}, \bibinfo{person}{Angelina McMillan-Major}, {and}
  \bibinfo{person}{Shmargaret Shmitchell}.} \bibinfo{year}{2021}\natexlab{}.
\newblock \showarticletitle{On the Dangers of Stochastic Parrots: Can Language
  Models Be Too Big?}. In \bibinfo{booktitle}{\emph{Proc. of FAccT '21}}.
\newblock


\bibitem[Benkler et~al\mbox{.}(2018)]%
        {benkler2018network}
\bibfield{author}{\bibinfo{person}{Yochai Benkler}, \bibinfo{person}{Robert
  Faris}, {and} \bibinfo{person}{Hal Roberts}.}
  \bibinfo{year}{2018}\natexlab{}.
\newblock \bibinfo{booktitle}{\emph{Network propaganda: Manipulation,
  disinformation, and radicalization in American politics}}.
\newblock \bibinfo{publisher}{Oxford University Press}.
\newblock


\bibitem[Bollenbacher et~al\mbox{.}(2021)]%
        {bollenbacher2021challenges}
\bibfield{author}{\bibinfo{person}{John Bollenbacher}, \bibinfo{person}{Diogo
  Pacheco}, \bibinfo{person}{Pik-Mai Hui}, \bibinfo{person}{Yong-Yeol Ahn},
  \bibinfo{person}{Alessandro Flammini}, {and} \bibinfo{person}{Filippo
  Menczer}.} \bibinfo{year}{2021}\natexlab{}.
\newblock \showarticletitle{On the challenges of predicting microscopic
  dynamics of online conversations}.
\newblock \bibinfo{journal}{\emph{Applied Network Science}}
  \bibinfo{volume}{6}, \bibinfo{number}{1} (\bibinfo{year}{2021}),
  \bibinfo{pages}{1--21}.
\newblock


\bibitem[Cantador et~al\mbox{.}(2020)]%
        {cantador2020exploiting}
\bibfield{author}{\bibinfo{person}{Iv{\'a}n Cantador},
  \bibinfo{person}{Mar{\'\i}a~E Cort{\'e}s-Cediel}, {and}
  \bibinfo{person}{Miriam Fern{\'a}ndez}.} \bibinfo{year}{2020}\natexlab{}.
\newblock \showarticletitle{Exploiting Open Data to analyze discussion and
  controversy in online citizen participation}.
\newblock \bibinfo{journal}{\emph{Information Processing \& Management}}
  \bibinfo{volume}{57}, \bibinfo{number}{5} (\bibinfo{year}{2020}),
  \bibinfo{pages}{102301}.
\newblock


\bibitem[Chen(2018)]%
        {chen2018spiral}
\bibfield{author}{\bibinfo{person}{Hsuan-Ting Chen}.}
  \bibinfo{year}{2018}\natexlab{}.
\newblock \showarticletitle{Spiral of silence on social media and the
  moderating role of disagreement and publicness in the network: Analyzing
  expressive and withdrawal behaviors}.
\newblock \bibinfo{journal}{\emph{New Media \& Society}} \bibinfo{volume}{20},
  \bibinfo{number}{10} (\bibinfo{year}{2018}), \bibinfo{pages}{3917--3936}.
\newblock


\bibitem[Choi et~al\mbox{.}(2015)]%
        {choi2015characterizing}
\bibfield{author}{\bibinfo{person}{Daejin Choi}, \bibinfo{person}{Jinyoung
  Han}, \bibinfo{person}{Taejoong Chung}, \bibinfo{person}{Yong-Yeol Ahn},
  \bibinfo{person}{Byung-Gon Chun}, {and} \bibinfo{person}{Ted~Taekyoung
  Kwon}.} \bibinfo{year}{2015}\natexlab{}.
\newblock \showarticletitle{Characterizing conversation patterns in reddit:
  From the perspectives of content properties and user participation
  behaviors}. In \bibinfo{booktitle}{\emph{Proceedings of the 2015 acm on
  conference on online social networks}}. \bibinfo{pages}{233--243}.
\newblock


\bibitem[Cinelli et~al\mbox{.}(2021)]%
        {cinelli2021echo}
\bibfield{author}{\bibinfo{person}{Matteo Cinelli}, \bibinfo{person}{Gianmarco
  De~Francisci Morales}, \bibinfo{person}{Alessandro Galeazzi},
  \bibinfo{person}{Walter Quattrociocchi}, {and} \bibinfo{person}{Michele
  Starnini}.} \bibinfo{year}{2021}\natexlab{}.
\newblock \showarticletitle{The echo chamber effect on social media}.
\newblock \bibinfo{journal}{\emph{Proceedings of the National Academy of
  Sciences}} \bibinfo{volume}{118}, \bibinfo{number}{9} (\bibinfo{year}{2021}).
\newblock


\bibitem[Cogan et~al\mbox{.}(2012)]%
        {cogan2012reconstruction}
\bibfield{author}{\bibinfo{person}{Peter Cogan}, \bibinfo{person}{Matthew
  Andrews}, \bibinfo{person}{Milan Bradonjic}, \bibinfo{person}{W~Sean
  Kennedy}, \bibinfo{person}{Alessandra Sala}, {and} \bibinfo{person}{Gabriel
  Tucci}.} \bibinfo{year}{2012}\natexlab{}.
\newblock \showarticletitle{Reconstruction and analysis of twitter conversation
  graphs}. In \bibinfo{booktitle}{\emph{Proceedings of the First ACM
  International Workshop on Hot Topics on Interdisciplinary Social Networks
  Research}}. \bibinfo{pages}{25--31}.
\newblock


\bibitem[Cohen(1989)]%
        {cohen1989deliberation}
\bibfield{author}{\bibinfo{person}{Joshua Cohen}.}
  \bibinfo{year}{1989}\natexlab{}.
\newblock \showarticletitle{Deliberation and Democratic Legitimacy}.
\newblock In \bibinfo{booktitle}{\emph{The Good Polity: Normative Analysis of
  the State}}, \bibfield{editor}{\bibinfo{person}{Alan Hamlin} {and}
  \bibinfo{person}{Phillip Petit}} (Eds.). \bibinfo{publisher}{Blackwell},
  \bibinfo{address}{New York}.
\newblock


\bibitem[Draws et~al\mbox{.}(2022)]%
        {draws}
\bibfield{author}{\bibinfo{person}{Tim Draws}, \bibinfo{person}{Oana Inel},
  \bibinfo{person}{Nava Tintarev}, \bibinfo{person}{Christian Baden}, {and}
  \bibinfo{person}{Benjamin Timmermans}.} \bibinfo{year}{2022}\natexlab{}.
\newblock \showarticletitle{Comprehensive Viewpoint Representations for a
  Deeper Understanding of User Interactions With Debated Topics}. In
  \bibinfo{booktitle}{\emph{ACM SIGIR Conference on Human Information
  Interaction and Retrieval}} (Regensburg, Germany)
  \emph{(\bibinfo{series}{CHIIR '22})}. \bibinfo{publisher}{Association for
  Computing Machinery}, \bibinfo{address}{New York, NY, USA},
  \bibinfo{pages}{135–145}.
\newblock
\showISBNx{9781450391863}
\urldef\tempurl%
\url{https://doi.org/10.1145/3498366.3505812}
\showDOI{\tempurl}


\bibitem[Dryzek(2009)]%
        {dryzek2009democratization}
\bibfield{author}{\bibinfo{person}{John~S. Dryzek}.}
  \bibinfo{year}{2009}\natexlab{}.
\newblock \showarticletitle{Democratization as Deliberative Capacity Building}.
\newblock \bibinfo{journal}{\emph{Comparative Political Studies}}
  \bibinfo{volume}{42}, \bibinfo{number}{11} (\bibinfo{year}{2009}).
\newblock
\urldef\tempurl%
\url{https://doi.org/10.1177/0010414009332129}
\showDOI{\tempurl}


\bibitem[Freelon et~al\mbox{.}(2020)]%
        {freelon2020false}
\bibfield{author}{\bibinfo{person}{Deen Freelon}, \bibinfo{person}{Alice
  Marwick}, {and} \bibinfo{person}{Daniel Kreiss}.}
  \bibinfo{year}{2020}\natexlab{}.
\newblock \showarticletitle{False equivalencies: Online activism from left to
  right}.
\newblock \bibinfo{journal}{\emph{Science}} \bibinfo{volume}{369},
  \bibinfo{number}{6508} (\bibinfo{year}{2020}), \bibinfo{pages}{1197--1201}.
\newblock


\bibitem[Garimella et~al\mbox{.}(2018)]%
        {garimella2018political}
\bibfield{author}{\bibinfo{person}{Kiran Garimella}, \bibinfo{person}{Gianmarco
  De~Francisci~Morales}, \bibinfo{person}{Aristides Gionis}, {and}
  \bibinfo{person}{Michael Mathioudakis}.} \bibinfo{year}{2018}\natexlab{}.
\newblock \showarticletitle{Political discourse on social media: Echo chambers,
  gatekeepers, and the price of bipartisanship}. In
  \bibinfo{booktitle}{\emph{Proceedings of the 2018 World Wide Web
  Conference}}. \bibinfo{pages}{913--922}.
\newblock


\bibitem[Gebru et~al\mbox{.}(2018)]%
        {timnit}
\bibfield{author}{\bibinfo{person}{Timnit Gebru}, \bibinfo{person}{Jamie
  Morgenstern}, \bibinfo{person}{Briana Vecchione},
  \bibinfo{person}{Jennifer~Wortman Vaughan}, \bibinfo{person}{Hanna~M.
  Wallach}, \bibinfo{person}{Hal~Daum{\'{e}} III}, {and} \bibinfo{person}{Kate
  Crawford}.} \bibinfo{year}{2018}\natexlab{}.
\newblock \showarticletitle{Datasheets for Datasets}.
\newblock \bibinfo{journal}{\emph{CoRR}}  \bibinfo{volume}{abs/1803.09010}
  (\bibinfo{year}{2018}).
\newblock


\bibitem[Glenski et~al\mbox{.}(2019)]%
        {glenski2019characterizing}
\bibfield{author}{\bibinfo{person}{Maria Glenski}, \bibinfo{person}{Emily
  Saldanha}, {and} \bibinfo{person}{Svitlana Volkova}.}
  \bibinfo{year}{2019}\natexlab{}.
\newblock \showarticletitle{Characterizing speed and scale of cryptocurrency
  discussion spread on reddit}. In \bibinfo{booktitle}{\emph{The World Wide Web
  Conference}}. \bibinfo{pages}{560--570}.
\newblock


\bibitem[Habermas(1984)]%
        {habermas1984theory}
\bibfield{author}{\bibinfo{person}{Jürgen Habermas}.}
  \bibinfo{year}{1984}\natexlab{}.
\newblock \bibinfo{booktitle}{\emph{The theory of communicative action}}.
\newblock \bibinfo{publisher}{Beacon Press}, \bibinfo{address}{Boston}.
\newblock
\showISBNx{0807015067 (v. 1)}


\bibitem[Hampton et~al\mbox{.}(2014)]%
        {hampton2014social}
\bibfield{author}{\bibinfo{person}{Keith~N Hampton}, \bibinfo{person}{Harrison
  Rainie}, \bibinfo{person}{Weixu Lu}, \bibinfo{person}{Maria Dwyer},
  \bibinfo{person}{Inyoung Shin}, {and} \bibinfo{person}{Kristen Purcell}.}
  \bibinfo{year}{2014}\natexlab{}.
\newblock \bibinfo{booktitle}{\emph{Social media and the `spiral of silence'}}.
\newblock \bibinfo{publisher}{PewResearchCenter}.
\newblock


\bibitem[Helberger(2019)]%
        {helberger}
\bibfield{author}{\bibinfo{person}{Natali Helberger}.}
  \bibinfo{year}{2019}\natexlab{}.
\newblock \showarticletitle{On the Democratic Role of News Recommenders}.
\newblock \bibinfo{journal}{\emph{Digital Journalism}} \bibinfo{volume}{7},
  \bibinfo{number}{8} (\bibinfo{year}{2019}), \bibinfo{pages}{993--1012}.
\newblock
\urldef\tempurl%
\url{https://doi.org/10.1080/21670811.2019.1623700}
\showDOI{\tempurl}
\showeprint{https://doi.org/10.1080/21670811.2019.1623700}


\bibitem[Joseph et~al\mbox{.}(2021)]%
        {joseph2021misalignment}
\bibfield{author}{\bibinfo{person}{Kenneth Joseph}, \bibinfo{person}{Sarah
  Shugars}, \bibinfo{person}{Ryan Gallagher}, \bibinfo{person}{Jon Green},
  \bibinfo{person}{Alexi~Quintana Mathé}, \bibinfo{person}{Zijian An}, {and}
  \bibinfo{person}{David Lazer}.} \bibinfo{year}{2021}\natexlab{}.
\newblock \showarticletitle{(Mis)alignment Between Stance Expressed in Social
  Media Data and Public Opinion Surveys}. In
  \bibinfo{booktitle}{\emph{Proceedings of the 2021 Conference on Empirical
  Methods in Natural Language Processing (EMNLP)}}.
\newblock


\bibitem[Kreiss(2021)]%
        {kreiss2021social}
\bibfield{author}{\bibinfo{person}{Daniel Kreiss}.}
  \bibinfo{year}{2021}\natexlab{}.
\newblock \bibinfo{title}{“Social Media and Democracy: The State of the
  Field, Prospects for Reform,” edited by Nathaniel Persily and Joshua A.
  Tucker}.
\newblock
\newblock


\bibitem[K{\"u}{\c{c}}{\"u}k and Can(2020)]%
        {kuccuk2020stance}
\bibfield{author}{\bibinfo{person}{Dilek K{\"u}{\c{c}}{\"u}k} {and}
  \bibinfo{person}{Fazli Can}.} \bibinfo{year}{2020}\natexlab{}.
\newblock \showarticletitle{Stance detection: A survey}.
\newblock \bibinfo{journal}{\emph{ACM Computing Surveys (CSUR)}}
  \bibinfo{volume}{53}, \bibinfo{number}{1} (\bibinfo{year}{2020}),
  \bibinfo{pages}{1--37}.
\newblock


\bibitem[Kumar et~al\mbox{.}(2010)]%
        {kumar2010dynamics}
\bibfield{author}{\bibinfo{person}{Ravi Kumar}, \bibinfo{person}{Mohammad
  Mahdian}, {and} \bibinfo{person}{Mary McGlohon}.}
  \bibinfo{year}{2010}\natexlab{}.
\newblock \showarticletitle{Dynamics of conversations}. In
  \bibinfo{booktitle}{\emph{Proceedings of the 16th ACM SIGKDD international
  conference on Knowledge discovery and data mining}}.
  \bibinfo{pages}{553--562}.
\newblock


\bibitem[Lee et~al\mbox{.}(2014)]%
        {lee2014social}
\bibfield{author}{\bibinfo{person}{Jae~Kook Lee}, \bibinfo{person}{Jihyang
  Choi}, \bibinfo{person}{Cheonsoo Kim}, {and} \bibinfo{person}{Yonghwan Kim}.}
  \bibinfo{year}{2014}\natexlab{}.
\newblock \showarticletitle{Social media, network heterogeneity, and opinion
  polarization}.
\newblock \bibinfo{journal}{\emph{Journal of communication}}
  \bibinfo{volume}{64}, \bibinfo{number}{4} (\bibinfo{year}{2014}),
  \bibinfo{pages}{702--722}.
\newblock


\bibitem[Loecherbach et~al\mbox{.}(2020)]%
        {Loecherbach}
\bibfield{author}{\bibinfo{person}{Felicia Loecherbach},
  \bibinfo{person}{Judith Moeller}, \bibinfo{person}{Damian Trilling}, {and}
  \bibinfo{person}{Wouter van Atteveldt}.} \bibinfo{year}{2020}\natexlab{}.
\newblock \showarticletitle{The Unified Framework of Media Diversity: A
  Systematic Literature Review}.
\newblock \bibinfo{journal}{\emph{Digital Journalism}} \bibinfo{volume}{8},
  \bibinfo{number}{5} (\bibinfo{year}{2020}), \bibinfo{pages}{605--642}.
\newblock
\urldef\tempurl%
\url{https://doi.org/10.1080/21670811.2020.1764374}
\showDOI{\tempurl}
\showeprint{https://doi.org/10.1080/21670811.2020.1764374}


\bibitem[Mansbridge(1999)]%
        {mansbridge1999everyday}
\bibfield{author}{\bibinfo{person}{Jane Mansbridge}.}
  \bibinfo{year}{1999}\natexlab{}.
\newblock \showarticletitle{Everyday talk in the deliberative system}.
\newblock In \bibinfo{booktitle}{\emph{Deliberative Politics: Essays on
  Democracy and Disagreement}}, \bibfield{editor}{\bibinfo{person}{Stephen
  Macedo}} (Ed.). \bibinfo{publisher}{Oxford University Press},
  \bibinfo{pages}{1--211}.
\newblock


\bibitem[Mansbridge(2015)]%
        {mansbridge2015minimalist}
\bibfield{author}{\bibinfo{person}{Jane Mansbridge}.}
  \bibinfo{year}{2015}\natexlab{}.
\newblock \showarticletitle{A minimalist definition of deliberation}.
\newblock \bibinfo{journal}{\emph{Deliberation and development: Rethinking the
  role of voice and collective action in unequal societies}}
  (\bibinfo{year}{2015}), \bibinfo{pages}{27--50}.
\newblock


\bibitem[Mercier and Landemore(2012)]%
        {mercier2012reasoning}
\bibfield{author}{\bibinfo{person}{Hugo Mercier} {and}
  \bibinfo{person}{Hélène Landemore}.} \bibinfo{year}{2012}\natexlab{}.
\newblock \showarticletitle{Reasoning is for arguing: Understanding the
  successes and failures of deliberation}.
\newblock \bibinfo{journal}{\emph{Political Psychology}} \bibinfo{volume}{33},
  \bibinfo{number}{2} (\bibinfo{year}{2012}), \bibinfo{pages}{243--258}.
\newblock


\bibitem[Nguyen et~al\mbox{.}(2020)]%
        {bertweet}
\bibfield{author}{\bibinfo{person}{Dat~Quoc Nguyen}, \bibinfo{person}{Thanh
  Vu}, {and} \bibinfo{person}{Anh~Tuan Nguyen}.}
  \bibinfo{year}{2020}\natexlab{}.
\newblock \showarticletitle{{BERTweet: A pre-trained language model for English
  Tweets}}. In \bibinfo{booktitle}{\emph{Proceedings of the 2020 Conference on
  Empirical Methods in Natural Language Processing: System Demonstrations}}.
  \bibinfo{pages}{9--14}.
\newblock


\bibitem[Nishi et~al\mbox{.}(2016)]%
        {nishi2016reply}
\bibfield{author}{\bibinfo{person}{Ryosuke Nishi}, \bibinfo{person}{Taro
  Takaguchi}, \bibinfo{person}{Keigo Oka}, \bibinfo{person}{Takanori Maehara},
  \bibinfo{person}{Masashi Toyoda}, \bibinfo{person}{Ken-ichi Kawarabayashi},
  {and} \bibinfo{person}{Naoki Masuda}.} \bibinfo{year}{2016}\natexlab{}.
\newblock \showarticletitle{Reply trees in twitter: data analysis and branching
  process models}.
\newblock \bibinfo{journal}{\emph{Social Network Analysis and Mining}}
  \bibinfo{volume}{6}, \bibinfo{number}{1} (\bibinfo{year}{2016}),
  \bibinfo{pages}{26}.
\newblock


\bibitem[Nozza et~al\mbox{.}(2020)]%
        {nozza2020mask}
\bibfield{author}{\bibinfo{person}{Debora Nozza}, \bibinfo{person}{Federico
  Bianchi}, {and} \bibinfo{person}{Dirk Hovy}.}
  \bibinfo{year}{2020}\natexlab{}.
\newblock \showarticletitle{What the [mask]? making sense of language-specific
  BERT models}.
\newblock \bibinfo{journal}{\emph{arXiv preprint arXiv:2003.02912}}
  (\bibinfo{year}{2020}).
\newblock


\bibitem[Rogers et~al\mbox{.}(2020)]%
        {rogers-etal-2020-primer}
\bibfield{author}{\bibinfo{person}{Anna Rogers}, \bibinfo{person}{Olga
  Kovaleva}, {and} \bibinfo{person}{Anna Rumshisky}.}
  \bibinfo{year}{2020}\natexlab{}.
\newblock \showarticletitle{A Primer in {BERT}ology: What We Know About How
  {BERT} Works}.
\newblock \bibinfo{journal}{\emph{Transactions of the Association for
  Computational Linguistics}}  \bibinfo{volume}{8} (\bibinfo{year}{2020}),
  \bibinfo{pages}{842--866}.
\newblock
\urldef\tempurl%
\url{https://doi.org/10.1162/tacl_a_00349}
\showDOI{\tempurl}


\bibitem[Schaefer and Stede(2021)]%
        {schaefer2021argument}
\bibfield{author}{\bibinfo{person}{Robin Schaefer} {and}
  \bibinfo{person}{Manfred Stede}.} \bibinfo{year}{2021}\natexlab{}.
\newblock \showarticletitle{Argument Mining on Twitter: A survey}.
\newblock \bibinfo{journal}{\emph{it-Information Technology}}
  \bibinfo{volume}{63}, \bibinfo{number}{1} (\bibinfo{year}{2021}),
  \bibinfo{pages}{45--58}.
\newblock


\bibitem[Sen et~al\mbox{.}(2020)]%
        {sen2020reliability}
\bibfield{author}{\bibinfo{person}{Indira Sen}, \bibinfo{person}{Fabian
  Fl{\"o}ck}, {and} \bibinfo{person}{Claudia Wagner}.}
  \bibinfo{year}{2020}\natexlab{}.
\newblock \showarticletitle{On the reliability and validity of detecting
  approval of political actors in tweets}. In
  \bibinfo{booktitle}{\emph{Proceedings of the 2020 Conference on Empirical
  Methods in Natural Language Processing (EMNLP)}}.
  \bibinfo{pages}{1413--1426}.
\newblock


\bibitem[Shugars and Beauchamp(2019)]%
        {shugars2019keep}
\bibfield{author}{\bibinfo{person}{Sarah Shugars} {and}
  \bibinfo{person}{Nicholas Beauchamp}.} \bibinfo{year}{2019}\natexlab{}.
\newblock \showarticletitle{Why keep arguing? Predicting engagement in
  political conversations online}.
\newblock \bibinfo{journal}{\emph{Sage Open}} \bibinfo{volume}{9},
  \bibinfo{number}{1} (\bibinfo{year}{2019}),
  \bibinfo{pages}{2158244019828850}.
\newblock


\bibitem[Sunstein and Sunstein(2018)]%
        {sunstein2018republic}
\bibfield{author}{\bibinfo{person}{Cass Sunstein} {and} \bibinfo{person}{Cass~R
  Sunstein}.} \bibinfo{year}{2018}\natexlab{}.
\newblock \bibinfo{booktitle}{\emph{\# Republic}}.
\newblock \bibinfo{publisher}{Princeton university press}.
\newblock


\bibitem[Tromble(2018)]%
        {tromble2018thanks}
\bibfield{author}{\bibinfo{person}{Rebekah Tromble}.}
  \bibinfo{year}{2018}\natexlab{}.
\newblock \showarticletitle{Thanks for (actually) responding! How citizen
  demand shapes politicians’ interactive practices on Twitter}.
\newblock \bibinfo{journal}{\emph{New media \& society}} \bibinfo{volume}{20},
  \bibinfo{number}{2} (\bibinfo{year}{2018}), \bibinfo{pages}{676--697}.
\newblock


\bibitem[Tromble and Meffert(2016)]%
        {tromble2016life}
\bibfield{author}{\bibinfo{person}{Rebekah Tromble} {and}
  \bibinfo{person}{Michael Meffert}.} \bibinfo{year}{2016}\natexlab{}.
\newblock \showarticletitle{The life and death of frames: Dynamics of media
  frame duration}.
\newblock \bibinfo{journal}{\emph{International Journal of Communication}}
  \bibinfo{volume}{10} (\bibinfo{year}{2016}), \bibinfo{pages}{23}.
\newblock


\bibitem[Tromble and Wouters(2015)]%
        {tromble2015are}
\bibfield{author}{\bibinfo{person}{Rebekah Tromble} {and}
  \bibinfo{person}{Miriam Wouters}.} \bibinfo{year}{2015}\natexlab{}.
\newblock \showarticletitle{Are We Talking \textit{with} or \textit{past} One
  Another? {E}xamining Transnational Political Discourse across
  {W}estern-{M}uslim "Divides"}.
\newblock \bibinfo{journal}{\emph{International Studies Quarterly}}
  \bibinfo{volume}{59}, \bibinfo{number}{2} (\bibinfo{year}{2015}),
  \bibinfo{pages}{373--386}.
\newblock
\urldef\tempurl%
\url{https://doi.org/10.1111/isqu.12167}
\showDOI{\tempurl}


\bibitem[Vrijenhoek et~al\mbox{.}(2021)]%
        {vrijenhoek2021recommenders}
\bibfield{author}{\bibinfo{person}{Sanne Vrijenhoek}, \bibinfo{person}{Mesut
  Kaya}, \bibinfo{person}{Nadia Metoui}, \bibinfo{person}{Judith M{\"o}ller},
  \bibinfo{person}{Daan Odijk}, {and} \bibinfo{person}{Natali Helberger}.}
  \bibinfo{year}{2021}\natexlab{}.
\newblock \showarticletitle{Recommenders with a mission: assessing diversity in
  news recommendations}. In \bibinfo{booktitle}{\emph{Proceedings of the 2021
  Conference on Human Information Interaction and Retrieval}}.
  \bibinfo{pages}{173--183}.
\newblock


\bibitem[Zeng et~al\mbox{.}(2018)]%
        {zeng2018microblog}
\bibfield{author}{\bibinfo{person}{Xingshan Zeng}, \bibinfo{person}{Jing Li},
  \bibinfo{person}{Lu Wang}, \bibinfo{person}{Nicholas Beauchamp},
  \bibinfo{person}{Sarah Shugars}, {and} \bibinfo{person}{Kam-Fai Wong}.}
  \bibinfo{year}{2018}\natexlab{}.
\newblock \showarticletitle{Microblog conversation recommendation via joint
  modeling of topics and discourse}. In \bibinfo{booktitle}{\emph{Conf. of the
  North American Chapter of the Association for Computational Linguistics:
  Human Language Technologies}}. \bibinfo{pages}{375--385}.
\newblock


\end{thebibliography}

\end{document}